# A comprehensive study on Frequent Pattern Mining and Clustering categories for topic detection in Persian text stream

Elnaz Zafarani-Moattar[1], Mohammad Reza Kangavari [*, 2], Amir Masoud Rahmani [1, 3]

[1] Department of Computer Engineering, Science and Research Branch, Islamic Azad University, Tehran, Iran
[2] Department of Computer Engineering, Iran University of Science and Technology, Tehran, Iran
[3] Future Technology Research Center, National Yunlin University of Science and Technology, 123 University Road, Section 3, Douliou, Yunlin 64002, Taiwan

[*] Corresponding Author: kangavari@iust.ac.ir

## Abstract

Topic detection in social networks is extracting and summarizing frequent posts about an event or subject. Topic detection is a complex process and depends on language because it somehow needs to analyze text. There have been few studies on topic detection in Persian, and the existing algorithms are not remarkable. Therefore, we aimed to study topic detection in Persian. The objectives of this study are: 1) to conduct an extensive study on the best algorithms for topic detection, 2) to identify necessary adaptations to make these algorithms suitable for the Persian language, and 3) to evaluate their performance on Persian social network texts. To achieve these objectives, we have formulated two research questions: First, considering the lack of research in Persian, what modifications should be made to existing frameworks, especially those developed in English, to make them compatible with Persian? Second, how do these algorithms perform, and which one is superior? There are various topic detection methods that can be categorized into different categories. Frequent pattern and clustering are selected for this research, and a hybrid of both is proposed as a new category. Then, ten methods from these three categories are selected. All of them are re-implemented from scratch, changed, and adapted with Persian. These ten methods encompass different types of topic detection methods and have shown good performance in English. The text of Persian social network posts is used as the dataset. Additionally, a new multiclass evaluation criterion, called FS, is used in this paper for the first time in the field of topic detection. Approximately 1.4 billion tokens are processed during experiments. The results indicate that if we are searching for keyword-topics that are easily understandable by humans, the hybrid category is better. However, if the aim is to cluster posts for further analysis, the frequent pattern category is more suitable.

## 1 Introduction

Topic detection (TD) is the process of extracting and summarizing frequent posts about an event or subject. TD is an important subject in computer science due to the increasing volume of posts on social media. The majority of current TD systems address English, while a great share of information is available in other languages. TD depends on language, and methods available in one language cannot be directly applied to other languages. Thus, the methods should be either developed specifically for each language or adapted from other languages with some modifications [1]. For instance, methods developed for English are used in other languages like French or Spanish with minor or significant changes.

Topic detection in Persian text streams presents unique challenges. Persian, being a rich and complex language, poses difficulties in the analysis and interpretation of textual data. Some of the challenges faced in topic detection of Persian text streams include:

1. Morphological Complexity: Persian text exhibits complex morphological features, such as affixes and word variations. The morphological differences between Persian and English are quite significant. Persian is an Indo-European language while English is a Germanic language. One key difference lies in the structure of the words. Persian is an inflected language, meaning that it uses various prefixes, suffixes, and infixes to convey grammatical information. English, on the other hand, is mostly uninflected, relying on word order and function words for grammatical information. Another example of morphological difference between Persian and English is in the formation of plurals. In Persian, plurals are formed by adding specific suffixes to the singular form of a noun. The rules for plural formation are complex and often irregular. In English, plurals are mostly formed by adding "-s" or "-es" to the singular form, with a few exceptions such as "ox" becoming "oxen". The plural formation in English is generally more straightforward compared to Persian. These variations make it challenging to identify and group similar words, affecting the accuracy of preprocessing and postprocessing algorithms.
2. Lack of Resources: Compared to widely studied languages like English, Persian lacks comprehensive linguistic resources and annotated datasets. This scarcity hinders the development and evaluation of topic detection models specific to Persian text streams. Persian is a low-resource language [2] not only in methods but also in datasets. Currently, there are no publicly available Persian datasets for TD [3], and only a few papers exist on Persian TD.
3. Contextual Ambiguity: Persian text often contains ambiguous words or phrases that require a deep understanding of the context to determine the intended meaning. This ambiguity adds complexity to topic detection, as misinterpretation of words can lead to inaccurate topic assignments.

The main motivation for this research started when we needed a topic detector for the Persian language for a real-world application, and we decided to implement it. After many searches, only a few articles were found in Persian e.g. [3] [4] [5]. In other words, the methods, proposed or examined in the Persian language, are very limited. Therefore, we decided to change some of the top methods in this field (in other languages) and adapt them to Persian to use the best of them in our practical project. This fact has become a motivation for this research in order to study 10 methods in three categories in Persian. All methods are modified and improved to adapt Persian. As we said, Persian is a low-resource language [2]; therefore, we encountered several problems in implementing these methods. For example, preprocessing tools for the Persian language were not suitable. Hence, new tools have been developed for Persian. These tools not only should be efficient, but also they should be fast enough to overcome the huge number of posts arriving in the data stream [6]. Existing tools were not real-time in processing the huge number of posts in the data stream. Another example was background knowledge of the Persian language. We had to spend about 40 days preparing this background knowledge from Persian Wikipedia articles by processing approximately 1.4 billion tokens.

Posts collected from Telegram social media public channels are used as a dataset in this research. There are two differences between the dataset of this research and other researches. First, almost all those datasets are collected from Twitter, and their language is English but this dataset is collected from Telegram, and the language is Persian. Telegram is selected because it is very popular in Iran and most Iranian people use Telegram. The second difference is in the method of collecting the dataset. In similar tasks, the dataset is usually collected through a query, which is based on one or more keywords. In contrast, in this research, all messages sent in the public channels of Telegram are considered. In other words, the entire data flow is taken into consideration, rather than just a selected portion of the messages determined by one or more keywords. For this reason, a special tool for data collection has been developed in the ComInSys Laboratory for the first time.

Summing up the above-mentioned points, the main contributions of this study can be summarized as follows:

- A comprehensive study on topic detection methods. Ten methods have been studied in this paper. We have implemented all of them and changed them to adapt them for Persian. This provides a broad understanding of the methods and their capabilities. No previous research has compared this many methods for Persian.
- Utilizing a very large dataset. The dataset of this study is prepared by processing approximately 1.4 billion tokens. This helps us to get more accurate results.
- Comparing categories and considering a new category for comparison. This study considers and compares not only methods but also categories. A new hybrid category is proposed and compared to other categories, providing insights into the capabilities of various categories.
- Focus on the Persian language. Persian is considered a low-resource language [2], which makes it more challenging to work with compared to English due to the lack of tools and implementation requirements. This work is an advancement in Persian language processing. We have implemented and adapted all of these methods for Persian.

In section 2, we will discuss the different classifications of methods in the field of TD. Section 3 will cover the problem definition. The implementation details of FPM-based, clustering-based, and hybrid methods can be found in sections 4, 5, and 6, respectively. Details of the experiments and their comparison results are provided in section 7. Finally, section 8 is devoted to the conclusion.

## 2   Different categories for topic detection methods

Many methods have been provided for topic detection, such as [7] [8] [9] [10]. These methods can be classified into different categories. Some of them are provided in the following. The methods in TDT[1] field [11], are divided into two categories in traditional media: document-pivot and feature-pivot. All the methods implemented in this paper fall under feature–pivot category. In 2013, Farzindar et.al [12] published an article entitled "A survey of techniques for event detection in Twitter". They categorized detection methods to three categories: unsupervised, supervised, and hybrid. The methods implemented in this article are located in the unsupervised and hybrid categories. According to this article, any event can be categorized into two categories: specified and unspecified. The events collected by our collector system are unspecified, which is one of the advantages of this paper. In 2017, Hassan et.al [13], in their paper entitled "A surrey on real-time event detection from the Twitter data stream'' divided methods of event detection into four categories: term-interestingness, topic-modelling,

---
[1] Topic Detection & Tracking

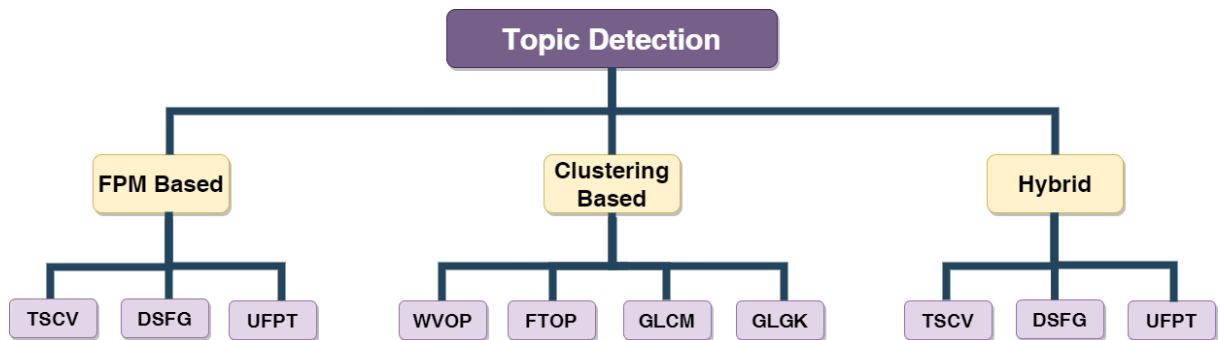

Figure 1: Topic detection methods implemented in this paper and their classification

incremental–clustering, and miscellaneous. The methods implemented in this paper are located in the term-interesting and incremental–clustering categories. Incremental clustering is needed for topic detection in data stream because each cluster represents a topic, and as new topics may emerge or an existing topic may fade, the number of topics (clusters) can vary over time. In 2018, Rania Ibrahim et al. [14], in their paper entitled "Tools and approaches for topic detection from Twitter streams: survey'' divided topic detection methods into five categories: Clustering, Frequent Pattern Mining, Exemplar-based, Matrix Factorization and Probabilistic Models. Current article focuses on the clustering and frequent pattern mining categories and the combination of these two methods. In 2021, Mottaghinia et al. [15] published a paper entitled "A survey on real-time event detection from the Twitter data stream". In this paper, the TD methods are classified into four categories: classification-based, clustering-based, topic-modelling-based and FCA-based. The methods implemented in our work fall into the clustering-based category. In 2021, Asgari-Chenaghlu et al. [16] in their paper entitled "Topic Detection and Tracking Techniques on Twitter: A Systematic Review" added a new category named deep-learning-based methods to the previous categories. All the clustering-based methods implemented in our work, utilize embedding methods and therefore, blong to this new category.

## 3    Problem Definition

Based on the different classifications mentioned in section 2, the Frequent Pattern (FP) category is one of the most important categories in the field of TDT. In this category, words are considered as items and frequent pattern mining algorithms are used to find topics. Another important category is the clustering (CL). This category is introduced after introducing embedding methods because measuring the distance between words and documents is essential in clustering, which is provided by word embedding. From our point of view, another category can be added to these categories, which is hybrid of frequent pattern and clustering (FPCL). We believe this because there are some methods in which, instead of using words of documents for clustering, frequent patterns of documents are extracted then these frequent patterns or their related documents are used for clustering.

Comparing the efficiency of methods from these three categories in the Persian data stream is one of our goals because there is a lack of researches in Persian that compares many methods together. We implemented 10 methods from scratch and modified them to adapt to Persian. These methods include three frequent-pattern-based methods, four clustering-based methods, and three hybrid methods. Figure 1 illustrates the topic detection methods implemented in this paper and their categories. All these 10 methods have been evaluated on the Sep_TD_Tel01 dataset [17], which is extracted from Telegram public channels in Persian. Telegram is very popular in Iran, which is why it was selected.

The implemented methods in this paper and their categories are given in Table 1. These 10 methods are classified into three categories: frequent pattern mining (FP), clustering (CL), and their combination as the hybrid model (FPCL). A four-letter standard is used for the naming the methods. In clustering methods, the first two letters represent the embedding method, and the subsequent two letters indicate the clustering method. In hybrid methods, the first two letters represent the frequent pattern mining method, and the next two letters show the clustering method.

Table 1: The implemented methods in this paper with their categories

| Row | Method | Category | Embedding Method | Frequent Pattern Mining Method | Clustering Method |
|---|---|---|---|---|---|
| 1 | TSCV | Frequent Pattern Mining (FP) | × | Term Selection + Co-occurrence Vector | × |
| 2 | DSFG | | × | Dynamic Support FP-Growth | × |
| 3 | UFPT | | × | Utility + FP-Tree | × |
| 4 | WVOP | Clustering (CL) | Word2Vec | × | OPTICS |
| 5 | FTOP | | fastText | × | OPTICS |
| 6 | GLCM | | GloVe | × | C-Means |
| 7 | GLGK | | GloVe | × | Gustafson-Kessel |
| 8 | SGJP | Hybrid (FPCL) | × | Segmentation | Jarvis-Patrick |
| 9 | CATT | | × | CIMAWA + AGF | TDA + TCTR |
| 10 | FHKN | | × | FP-Growth + HUP | K-means + Newman |

# 4  Frequent Pattern Mining-based Methods

Frequent pattern mining-based methods select patterns with high frequency in posts to detect the topic. In other words, a number of patterns that have high frequency in the whole posts are mined as the topic of that set of posts. Three frequent pattern mining-based methods are implemented in this paper. This section will explain how to implement them. In the first method, Wikipedia background knowledge is used to select terms. In the second method, dynamic support values are applied in the FP-Growth method. In the third method, the concept of utility is used to select patterns.

## 4-1  TSCV

We will study the first frequent pattern mining based method in this section. This method utilizes the background knowledge of Persian Wikipedia to select terms. The TSCV method has been implemented based on [18], with modifications made to it as described in steps 1 and 2. Additionally, this method is specified for the Persian language. The method consists of three phases: term selection, co-occurrence vector formation and post-processing as follows:

1. Term Selection: In this phase, k top-terms are selected among all the words within the window. Two corpora are used for this purpose. The first corpus, denoted as corpus$_{new}$ in eq. (1), is the new corpus which is being studied. In our implementation, the posts in each window are used as corpus$_{new}$. The second corpus, denoted as corpus$_{ref}$ in eq. (1), plays the role of background knowledge. In our implementation, the Wikipedia corpus is used as background knowledge. This corpus includes more than 3.2 million documents and more than 712 million tokens. Finally, the score of each word is calculated as follows:

$$score(w) = \frac{P(w|corpus_{new})}{P(w|corpus_{ref})} \tag{1}$$

Once the score for each word is calculated, the list of words is sorted and k top words are selected as top-terms and stored in the T list. Here, k is a tunable parameter. The tunable parameters and their corresponding value-setting tests are presented in sections 7-6.

2. Co-occurrence-vector formation: Actually, this phase is the core of this method. In the first step, a vector is created for each word in T with a length equal to the number of documents. These vectors contain binary values. The value in the index $i$ determines the presence (1) or the absence (0) of the word in the $i^{th}$ document. In the next step, an empty list is defined for the topics set. Then a loop is started to traverse the words in the set T. Within this loop, at each step, the word $t$ is added to the variable $S$, and the vector of word $t$ is added to the variable $D_s$. Subsequently, the algorithm searches for a word within the list T (excluding $t$), whose vector is most similar to the vector of the word $t$ (based on cosine similarity criteria), and stores it in $\hat{t}$ (best matching term). Then, $\theta$ is obtained as follows:

$$\theta(S) = 1 - \frac{1}{1 + exp((|S| - b)/c)} \tag{2}$$

where b = 5 and c = 2 are constants and the set $S$ has a variable length as new words may be added at each iteration of the loop. Once $\theta$ is calculated, the similarity between two vectors $D_S$ and $D_{\hat{t}}$ is compared with $\theta$. if the similarity is higher than $\theta$, then the word $\hat{t}$ is added to set $S$, and vector $D_{\hat{t}}$ is added to the vector $D_S$ (the values of two vectors are summed correspondingly). Then, the word $\hat{t}$ is removed from set T.

Inspecting the primary results of this algorithm, we notice that the vector $D_S$ may be filled with small non-zero values after a while, especially, if a high-frequency term is added to $S$. It is expected that high-frequency words are added to $S$, but this problem may lead to adding words that are similar to only few words in $S$. To prevent this, at each iteration of the loop, after updating the vector $D_S$, any value of this vector that is less than $\frac{|S|}{2}$ is replaced with "0". Then, the next iteration of the loop is executed. This process continues until the obtained term-similarity is less than $\theta$. In this case, the set $S$ is added to the topics set, and the first iteration of the loop is finished. Afterward, all of these processes are repeated for the next term in set T until all members of this set are traversed, and this set becomes empty.

3. Post-processing: In the final step of this algorithm, the topics set is traversed to remove duplicate topics.

## 4-2 DSFG

The DSFG method is implemented based on [19]. A prominent feature of this method is the implementation of FP-Growth with dynamic "min support". The FP-Growth algorithm is one of the most commonly used methods for frequent pattern mining. In this algorithm, an FP-Tree is formed to detect frequent patterns. A parameter called min support is used to prune this tree. Determining the value of min support is the main problem of the FP-Growth algorithm because it is not only difficult to determine its optimal value in a window but also its optimal value changes in each window. In the dynamic support method, the support value in calculated differently depending on the window size. The windows are divided into two classes: small and large, based on their size. In our implementation, a window is considered small when the number of posts in that window is less than one-third of the average number of posts in the preceding few windows. Equation (3) is employed to calculate support for a small windows, while eqation (4) is used for large windows.

$$Sup^t = \underset{w_i \in B^t}{\text{avg}} \{TF(w_i)\} \times (2 \times \underset{w_i \in B^t}{\text{median}} \{TF(w_i)\}) \quad (3)$$

$$Sup^t = \underset{w_i \in B^t}{\text{avg}} \{TF(w_i)\} \times \underset{w_i \in B^t}{\text{median}} \{TF(w_i)\} \quad (4)$$

where $B^t$ represents the current window (time t) and $TF(w_i)$ represents the term frequency of word $w_i$.

### 4-3  UFPT

The third method of frequent pattern mining is discussed in this section. This method is implemented based on [20] by applying following modifications. UFPT utilizes a concept called utility to extract frequent patterns. The utility is a concept introduced by Choi and Park [20] to detect emerging topics in the Twitter stream. Their method is built upon concepts such as DB table, transaction table, and item set, which were introduced by Lio & Qu [21]. Although these concepts were originally introduced in [21] in the context of databases, they have been adapted for the data stream and social network in [20]. In our implementation, we have adapted these concepts for the Telegram data stream. In our implementation, each post is considered as a transaction and the words of each post are treated as items. This allows us to detect emerging topics in social network posts using High Utility Pattern Mining (HUPM) such that both frequency and utility play a role in selecting patterns. The utility of each word is calculated by multiplying its internal and external utility. The internal utility is defined as the frequency of that word in a post. The external utility is used to identify an emerging word with high frequency in the current window. The frequency difference between two consecutive batches of posts is taking into account when calculating the external utility. If this difference is positive, it indicates that the frequency of the word $w_i$ is currently increasing. Conversely, if the difference is negative, it indicates that this frequency is currently decreasing.

Next, the transaction utility (TU) for a post is obtained by summing the utilities of the words in the post. Finally, the transaction-weighted utility (TWU) is calculated for every word in a batch. $TWU(w_i)$ for a word $w_i$ is the sum of TUs for all posts that contain the word $w_i$. The purpose of calculating $TWU(w_i)$ is to filter out less important words by establishing a threshold called min-util. Words with a TWU value lower than min-util are then filtered out. Finally, topics are extracted using the HUPM algorithm [22] on the remaining words and then applying the TP-tree algorithm [23] on the output of HUPM.

## 5  Clustering-based Methods

Four clustering-based methods are implemented in this paper, and their implementation details are explained in this section. Clustering-based methods are those methods that apply clustering algorithms directly to the input text for topic detection. In these methods, the posts are placed in clusters based on their similarity, and each cluster is presented as a topic. The similarity of the posts is calculated by the distance between the embedding vectors of the input posts. Since Persian is a low-resource language, finding a suitable embedding was one of the main challenges in this research because most of the available embeddings had a high out-of-vocabulary rate. We had to choose the most suitable embeddings from the available options. Two distance metrics, Euclidean distance, and cosine distance, are used to calculate the distances of vectors in this paper. These two distance metrics are well-known metrics that are commonly used in topic detection. The cosine distance between documents $d_i$ and $d_j$ is obtained by the cosine similarity [24], as shown in equation (5):

$$dist_{cos}(\vec{d_i}, \vec{d_j}) = 1 - sim_{cos}(\vec{d_i}, \vec{d_j}) \tag{5}$$

## 5-1  WVOP

In this method, Word2Vec embedding is used to obtain the embedding vector of input posts, and the OPTICS algorithm is applied to cluster these vectors. The size of the Word2Vec vector is set to 200, and the cosine distance metric is used to calculate the distance between the vectors. The OPTICS algorithm is one of the density-based algorithms. These algorithms have two main advantages: 1) They are not limited to classes with a specific shape (e.g. spherical) and can identify classes with any shape. 2) There is no need for prior knowledge about the number of classes.

## 5-2  FTOP

In the second method of clustering-based methods, fastText is used to obtain the embedding vector of input posts, and the OPTICS algorithm is used to cluster these vectors. The size of the fastText vector is set to 200, and the cosine distance metric is used to calculate the distance between the vectors.

## 5-3  GLCM

The third clustering-based method is GLCM. This method uses GloVe for embedding and C-Means (CM) for clustering. Size of GloVe embedding vectors in this implementation is 200 and parameters m and epsilon are assumed 1.1 and 0.001 respectively. This method is implemented based on [25].

## 5-4  GLGK

The fourth method of clustering-based methods is GLGK which uses GloVe as embedding and Gustafson-Kessel (GK) as clustering algorithm. The GK algorithm uses $D_{GK}$ distance metric for clustering which is calculated as:

$$D_{GK}^2 = (x_k - V_i)^T A_i (x_k - V_i) \tag{6}$$

$$A_i = (\rho_i |C_i|)^{1/d} C_i^{-1} \tag{7}$$

$$C_i = \frac{\sum_{k=1}^{N} \mu_{ik}^m (x_k - V_i)^T (x_k - V_i)}{\sum_{k=1}^{N} \mu_{ik}^m} \tag{8}$$

where parameters m and epsilon are assumed 1.1 and 0.001 respectively and the size of GloVe embedding vectors is set to 200 in this implementation. This method is implemented based on [25].

## 6  Hybrid Methods (FPCL)

Hybrid methods are a combination of FPM (section 4) and clustering (section 5). Hybrid methods concentrate on clustering FPs instead of clustering posts. In this category, three papers based on the combination of FPM and clustering have been selected. All three articles are on Twitter and in English. We have implemented the methods of all three papers (with the mentioned modifications) on the Telegram Sep_TD_Tel01 dataset. These three methods are discussed in sections 6-1, 6-2, and 6-3.

## 6-1 SGJP

SGJP uses segmentation for frequent pattern mining and Jarvis-Patrick for clustering. Jarvis-Patrick is a graph-based clustering algorithm. Topic detection in this method is based on segmentation. Another feature of this method is to employ background knowledge in the topic detection process.

SGJP method, which is based on [26], consists of three main phases: post segmentation, topic segment detection and topic segment clustering. In "post segmentation" phase, input posts are divided into non-overlapping segments. The segmentation process gives this method the ability to select better and meaningful phrases for topics. These segments are consecutive words in posts. Segments seen in many posts are more likely to be named entities or meaningful phrases. Background knowledge has been used to extract meaningful phrases. Three corpora, Hamshahri newspaper, posts collected from Telegram and Wikipedia Persian articles, are used to produce background knowledge. Hamshahri and Telegram corpora are used to obtain n-gram probabilities, and the Wikipedia corpus is used to extract anchor texts and to calculate probabilities. In "post segmentation" phase, the stickiness function for each segment s is defined as C(s) in eq. (9). A criterion called SCP[1] is defined to calculate the coherence of the segment s. This criterion considers all possible binary segmentations.

$$C(s) = Len(s) \cdot e^{Q(s)} \cdot Sig(SCP(s)) \tag{9}$$

$$Len(s) = \begin{cases} \dfrac{|s|-1}{|s|} & , |s| > 1 \\ \dfrac{1}{3} & , |s| = 1 \end{cases} \tag{10}$$

$$SCP(s) = Log\left(\dfrac{\Pr(s)^2}{\dfrac{1}{n-1}\sum_{i=1}^{n-1}\Pr(w_1 \ldots w_i)\Pr(w_{i+1} \ldots w_n)}\right) \tag{11}$$

In our implementation, we considered $Q(s)$ in eq. (9) as the probability that the segment s appears as an anchor text in Wikipedia articles. $Sig(.)$ is a sigmoid function. The goal of using $Len(.)$ is to prefer longer segments to shorter segments. In eq. (11), Pr (.) is the prior probability. In our implementation, it is obtained from the n-gram probability for N>=2. Hamshahri newspaper and Telegram corpora have been used to calculate this probability. Approximately, 1.4 billion tokens are processed to prepare Persian anchor text and n-gram probabilities, taking about 40 days. In the "topic segment clustering" phase, a graph is constructed after extracting meaningful segments. Finally, clustering is performed on this graph using the Jarvis-Patrick algorithm resulting in clusters that consist of segments sharing a common subject.

## 6-2 CATT

This method uses CIMAWA and AGF to extract frequent patterns and a combination of TDA and TCTR algorithms to perform the clustering phase. This hybrid method is introduced by Benny et al. [27].

CATT uses a concept called CIMAWA, which mimics the human mental ability of word correlation. Based on a test with 6000 participants conducted by Steyvers et al. [28], a significant part of human word association is asymmetric. For instance, "canary" is more associated with "bird" than "bird" to "canary". In order to achieve word correlation rate, as a human perceives it, and also the ability to combine symmetric and asymmetric word correlation, a concept called CIMAWA is defined base on the co-occurrence of the words. The definition of co-occurrence in the original paper is somewhat vague and lacks clarity. For instance, it is unclear whether the order or the presence of additional words between x and y matters in determining co-occurrence. In our implementation, x and y are considered co-occurred if both of them are included in a post, regardless of their placement. CIMAWA can be calculated as:

$$CIMAWA(x,y) = \frac{Cooc(x,y)}{f(y)} + \delta \frac{Cooc(x,y)}{f(x)} \qquad (12)$$

where, in our implementation, Cooc(x,y) indicates number of posts in a window in which words x and y occur together. f(x) shows number of posts containing x and same for f(y). δ, which is called damping factor, is a number between 0 and 1. Its value is set to 0.5 in our implementation. Then, Association Gravity Force (AGF) is defined based on CIMAWA. The larger value for AGF, the bigger gravity between two frequent patterns, which means the occurrence of pattern y is strongly related to the occurrence of pattern x. In this method, the patterns are sorted by the value of AGF, and the top patterns are selected. Finally, clustering is performed on selected patterns using TDA and TCTR algorithms [27].

---
[1] Symmetric Conditional Probability

## 6-3 FHKN

FHKN is the third method in the hybrid category. In this method, HUPC is used to extract frequent patterns and KNN and Newman algorithms are used to cluster extracted patterns. HUPC framework over the microblog stream is proposed in [22]. HUPC method is a hybrid method, i.e., it divides topic detection problem into two sub-problems: pattern mining and clustering. This framework consists of four steps: 1) FP-Growth algorithm, 2) top-k HUP[1] mining, 3) HUP clustering, 4) postprocessing. Steps 1 and 2 tackle with pattern mining sub-problem. The clustering process includes steps 3 and 4.

FP-Growth algorithm [23][29] is used to generate frequent patterns. The minimum support value in FP-Growth is set to a very small value to generate more frequent patterns. Patterns generated by the FP-Growth are overlapped. HUP mining algorithm is used to select some of such overlapping patterns. "Utility" is defined as importance of a pattern. In this paper, the frequency of a word in the current window is considered as the utility of that word. HUP clustering is the next step. Two types of topics may occur in a window: coherent topics and emerging topics. The KNN is used to classify HUPs into coherent (existing) topics clusters based on its k nearest neighbors (i.e. similar patterns in the previous window), and the Newman modularity-based clustering is used to extract emerging (new) topics from the rest of the HUPs.

In our implementation, cosine similarity is used to measure the similarity between the current pattern and a coherent topics pattern (is used in the KNN algorithm), and the Jaccard coefficient is used to measure the similarity between two unclassified patterns in the current window (is used in the Newman algorithm). Note that measuring the pattern similarity based on the common words of patterns does not perform well. Therefore, in this paper, the similarity between two patterns is measured based on their related posts in a window. In the postprocessing step, the importance score is calculated for each word of each cluster. Score calculation details are similar to [22]. Some of the top-ranked words according to importance score are selected as the topic of the cluster.

---

[1] High Utility Pattern

# 7 Experiments

The Framework of the study is presented in Figure 2. This figure illustrates the process of producing background knowledge as well as the process of executing and evaluating methods for our experiments. First, the datasets used in this research are discussed in the following subsections. Then, details of evaluation, along with some samples, are given in the subsequent subsections. Many months were spent processing and preparing these datasets for use in this research. The FS criterion is used to evaluate the results. FS is a new multiclass – multicluster criterion which is being used to evaluate topic detection results for the first time in this research.

## 7-1 Dataset for Background Knowledge

In this study, a combination of three different datasets is used as background knowledge. These three datasets are the Hamshahri, Telegram and Wikipedia datasets. The Hamshahri dataset contains text from the Hamshahri newspaper, one of the oldest newspapers published in Iran in Persian. The Telegram dataset consists of Persian posts collected from the Telegram social network. This dataset includes all posts sent on Telegram's public channels, and its collection is unlimited, i.e., all posts are collected and no keyword or other constraint is used. The Wikipedia dataset includes Persian articles from Wikipedia. The details of these three datasets are provided in Table 2. The Hamshahri and Telegram datasets are used to obtain n-gram probabilities [30], while the Wikipedia dataset is used to extract anchor texts and calculate their probabilities. The number of n-grams and anchor texts are given in Table 3. It is worth mentioning that calculating the n-gram and anchor text probabilities took approximately 40 days.

Table 2: Details of datasets used as background knowledge

| Dataset | Number of Documents | Number of Tokens |
|---|---|---|
| Hamshahri | 83,326 | 67,437,494 |
| Telegram | 1,835,961 | 100,240,166 |
| Wikipedia | 3,254,132 | 712,472,799 |

Table 3: Number of n-grams and anchor texts extracted from datasets

| Property | Value |
|---|---|
| 1-gram | 99.603.396 |
| 2-gram | 94,305,722 |
| 3-gram | 89,158,246 |
| 4-gram | 84,110,988 |
| 5-gram | 79,176,683 |
| Anchor text | 9,570,984 |
| Distinct tokens | 2,107,360 |

## 7-2 Golden Standard Dataset for Topic Detection

The purpose of this paper is topic detection on social media text. The process is carried out on the Sep_TD_Tel01 dataset [17], which is in Persian. This dataset is collected from Telegram without any limitations, such as keywords, and therefore, it fully represents the data stream nature. Most scientific researches on topic detection are on the Twitter social network. Twitter's popularity stems from its provision of free API access to data for users [31][32]. In contrast, this research is on the Telegram social network because this social network is very popular in Iran. A message collector system was developed by ComInSys laboratory to access Telegram data. This program collects all messages from public channels and groups, of which it is a member. This is an advantage of our collector system, because this system collects all messages, while in the other systems, data collection is based on one or more keywords. As mentioned above, other researches use the Twitter API to collect data by filtering tweets with specific keywords. Therefore, collected datasets are biased on those keywords. In contrast, the dataset of this research is unbiased because all the messages are collected. For instance, during a specific time interval in Sep_TD_Tel01, two significant topics were discussed: "The death of Ayatollah Hashemi Rafsanjani" and "The fire in the Plasco building", which overlapped in their occurrence. Simultaneously, other topics such as the "inauguration of Trump" and the "resignation of Carlos Queiroz" were also being discussed. Now, if the collection is done by keyword #Plasco in the Twitter API, other topics in this period will be missed. This means that the dataset will not contain, for example, "The death of Ayatollah Hashemi Rafsanjani" or "inauguration of Trump", and the dataset will be limited to posts of #Plasco. In contrast, our collector system collects all posts in this period, which is an advantage for our dataset.

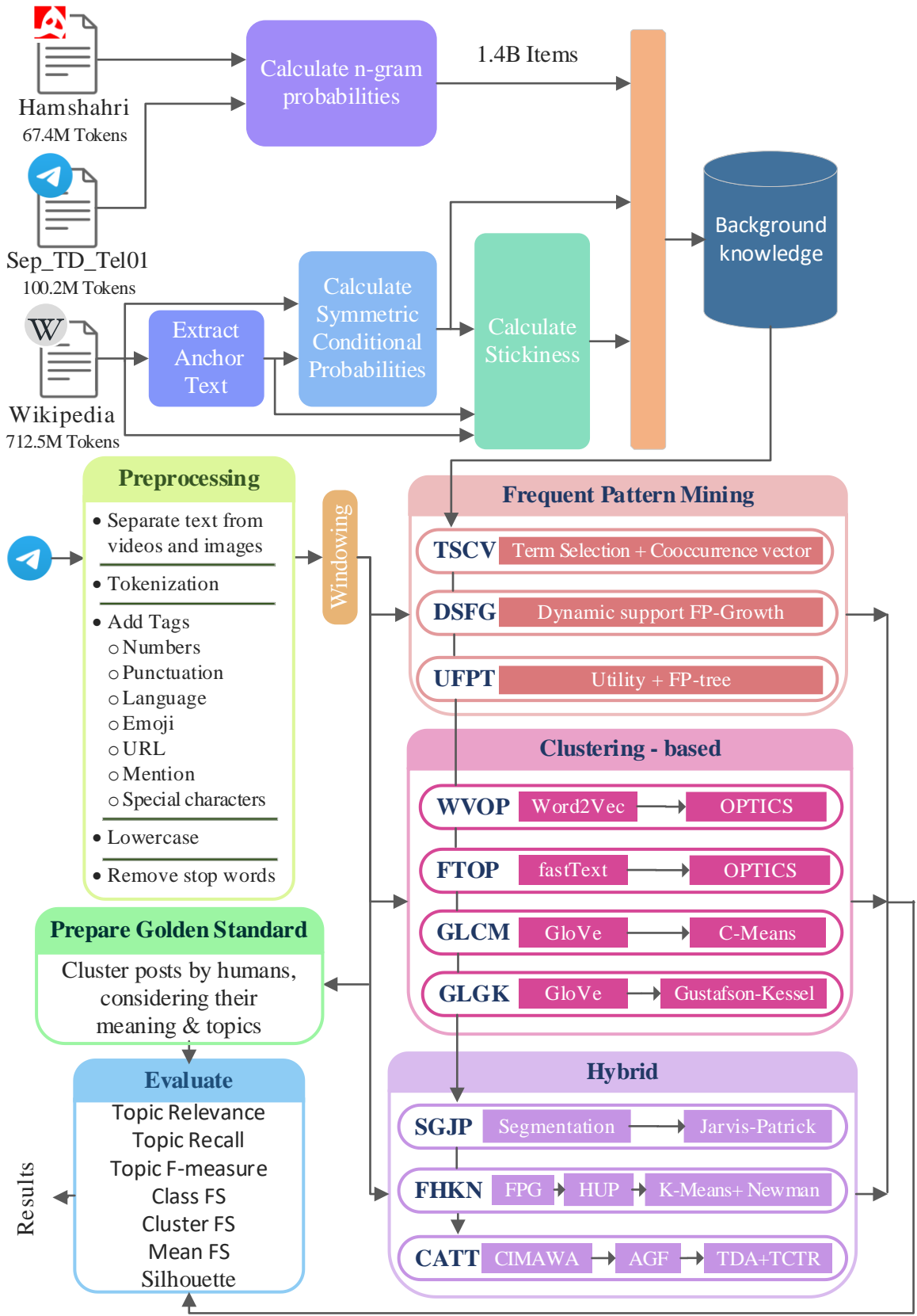

Figure 2: Framework of implemented methods and background knowledge extraction.

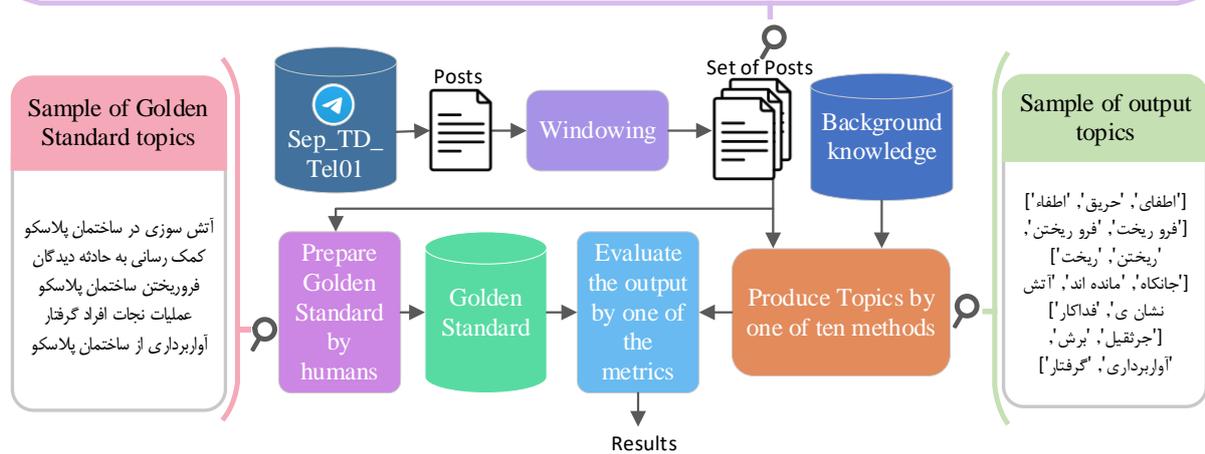

Figure 3: Evaluation framework. The process to produce golden standard as well as output of the methods and how to evaluate the methods in addition to samples of input posts, samples of output of methods and samples of golden standard

Since the testing platform of this paper is online topic detection from the data stream, windowing becomes necessary. Three windowing models are presented in [33] [34]: sliding window, damped window, and landmark window. In this study, the landmark window model has been employed on the telegram data stream.

As the volume of collected data is huge and humans cannot annotate all of it, a one month period from Jan 1, 2017, to Jan 31, 2017, was chosen to stablish the golden standard. This time frame was selected because two significant events (super topics) occurred in this period in Iran which are "The death of Ayatollah Hashemi Rafsanjani" and "The fire in the Plasco building". These topics remained hot for a long period of time. In addition to these two super topics, other topics such as the "inauguration of Trump" and the "resignation of Carlos Queiroz" exist during this period. The posts of this period of time are given to an expert for clustering them based on their topics. It is worth mentioning that hash tags are not considered for clustering, neither by humans nor by any of the methods. Human evaluators determine the topic of each post based on its meaning, and the methods do this based on the words of posts. It is possible for a post to be assigned to multiple clusters if it covers more than one topic. This applies to both the golden standard and the outputs of the methods. That is why FS evaluation metrics are used to evaluate outputs. 3 demonstrates the evaluation framework of the methods. This figure shows some samples of input posts and the corresponding outputs of the methods, as well as the corresponding golden standard samples. Additionally, the figure shows the process of producing the golden standard and evaluating the system.

### 7-3  Preprocessing

The published messages contain three media types: images, videos and text. As topics are detected from the text, only the textual content of the messages is considered. Tokenization is the first step in the preprocessing phase. A specialized tokenizer developed in the ComInSys lab is utilized for this purpose. The input text undergoes tokenization using this tokenizer, with additional tags being assigned to each token as necessary. These tags include tags for numbers, punctuation symbols, languages ("English word", "Persian word", etc.), emojis, hashtags, URLs, mentions and other special characters. Once the tokenization process is complete, tokens tagged with the "Persian word" identifier are selected and converted to lower case. Finally, stop words are eliminated.

### 7-4  Postprocessing

The postprocessing stage is responsible for creating the final titles for topics. Based on the word formation rules of Persian morphology, compound words (including verbs) are made from two or more words. These words sometimes have spaces between them. Placing space-separated but related words (compound words) together (in the title of topic) is one of the tasks of postprocessing. Additionally, this stage is responsible for assigning scores to selected words for the title in order to sort and select the best words.

The morphology of language has an impact on preprocessing and postprocessing stages. Persian has a rich morphology. Persian words can undergo various morphological changes depending on their grammatical role and context. For example, Persian has a much more extensive system of verb conjugation compared to English. Persian verbs have different forms based on tense, number, and person, as well as positive and negative statements, whereas English verbs have relatively few conjugation forms. The main changes in English verbs occur in the third person singular form and in the past tense. English morphology is generally less complex than Persian morphology.

### 7-5  Evaluation Metric

The evaluation metrics used in this research are presented in this section. These metrics are used not only to evaluate the implemented methods but also to compare the categories. This paper uses FS, which is a newly released multiclass - multicluster evaluation metric, for the first time.

In this research, we adopt two points of view to evaluate the methods. The first point of view focuses on evaluating the terms presented by the system as topics. This view concentrates on the quality of the terms generated by the system and their similarity to human opinions. In the second point of view, related posts of the topic are evaluated. The important thing in this viewpoint is the semantic relation of the posts. In other words, we want to determine if the system can accurately identify the semantic relation of the posts and if the system can cluster semantically related posts into an identical cluster. In a real system, significant topics are usually extracted from a mass of posts in the data stream and presented to the user, allowing them to understand the overall content of the posts in the data stream. Obviously, the quality of the words provided to the user as topics is important for better understanding the content. This is the aim of the first viewpoint. In contrast, usually, after extracting topics in the data stream, one or more important topics are explored by users. That is, the user investigates the posts of the topic. In this case, correctly identifying the posts related to that topic is important. This represents the second viewpoint in evaluating topics.

Seven metrics are used for evaluation in this study, six of which are in these two viewpoints. Three metrics, "Topic relevance", "Topic recall", and "Topic F-measure", fall into the former viewpoint, while, the three metrics, "Class FS", "Cluster FS", and "Mean FS", fall into the latter one. Additionally, the Silhouette metric is used independently to evaluate the clustering quality during parameter tuning.

"Topic relevance", "Topic recall", and "Topic F-measure" metrics are similar to precision, recall, and f-measure, but instead of being calculated over terms, they are calculated over topics. These metrics are presented in detail in [20].

In the second view, we use FS metrics. The FS metric is a new metric introduced in [35] for the first time. FS is a multiclass - multicluster evaluation metric which allows a sample to be placed in more than one class or cluster at a time. This is an important requirement to evaluate topics, as a post can be related to more than one topic. For example, a post announcing the closure of schools due to air pollution can be included in both the posts related to the closure of schools and the posts related to air pollution. As calculating FS is a little bit complicated, you can refer to [35] for details but briefly speaking, FS can be calculated as:

$$CluterFS(\Omega) = \frac{\sum_j FS(\omega_j) \sum_i Sr_{ij}}{\sum_i \sum_j Sr_{ij}} \quad (13)$$

$$ClassFS(C) = \frac{\sum_i FS(c_i) \sum_j Sc_{ij}}{\sum_i \sum_j Sc_{ij}} \quad (14)$$

where $FS(\omega_j)$ is the score of cluster $\omega_j$, $Sr_{ij}$ is the total score of the samples of cluster j assigned to class i. $FS(c_i)$ is the score of class $c_i$, and $Sc_{ij}$ is the total score of samples of class i assigned to cluster j. To unify these scores, the MeanFS metric is defined as follows:

$$MeanFS = \frac{1}{2}(ClassFS + ClusterFS) \quad (15)$$

As mentioned above, the silhouette metric, which is independent of these two viewpoints, is used to evaluate the quality of clustering during parameter setting.

### 7-6  Parameter Tuning

The tunable parameters of the methods are listed in Table 4. It is worth mentioning that there are many parameters in the methods, as described in sections 4, 5 and 6. These parameters can be classified into three types. Some of them been studied in the literature, and there accepted values are available for them. The value for this type of parameters are mentioned in sections 4, 5 and 6. The value of some others dependents on the posts of each window (such as the number of clusters in k-means). The value of these parameters needs to be determined in each window. The third type of parameters is the ones whose values should be examined to find the best values for our application and dataset. We refer to them as tunable parameters, and only this type of parameters are listed in Table 4.

For each method, a wide range of values is considered for tunable parameters, which are also listed in Table 4. Next, the method is run for each value of the parameter, and the best value is selected. If the method has more than one parameter, all combinations of values for all parameters are considered. This process was a time-consuming process which took hours to complete. Silhouette and FS are employed as criteria to select the optimal value for each parameter. In other words, the best value of each parameter p, is selected by the following formula:

$$p^* = \underset{p \in range(p)}{\mathrm{argmax}} \{Evaluate(Run\_method(Parameter = p))\} \tag{16}$$

Results of one of parameter tuning experiments are shown in Figure 4 as sample.

Table 4: Tunable parameters of different methods

| Row | Method | Parameters | Range of Values | Number of Experiments |
|---|---|---|---|---|
| 1 | TSCV | k | 20:10:90&100:50:300 | 117 |
| 2 | DSFG | - | - | - |
| 3 | UFPT | - | - | - |
| 4 | WVOP | MinPts | 2:30 | 261 |
| 5 | FTOP | MinPts | 2:30 | 261 |
| 6 | GLCM | - | - | - |
| 7 | GLGK | - | - | - |
| 8 | SGJP | h | 1:6 | 108000 |
| | | Threshold | 1:20 | |
| | | k | 10:10:100 | |
| | | k_min | 5:5:k-5 | |
| 9 | CATT | Rate | 5%:5:50% | 900 |
| | | Damp | 0.1:0.1:1 | |
| 10 | FHKN | - | - | - |
| sum | | | | 109539 |

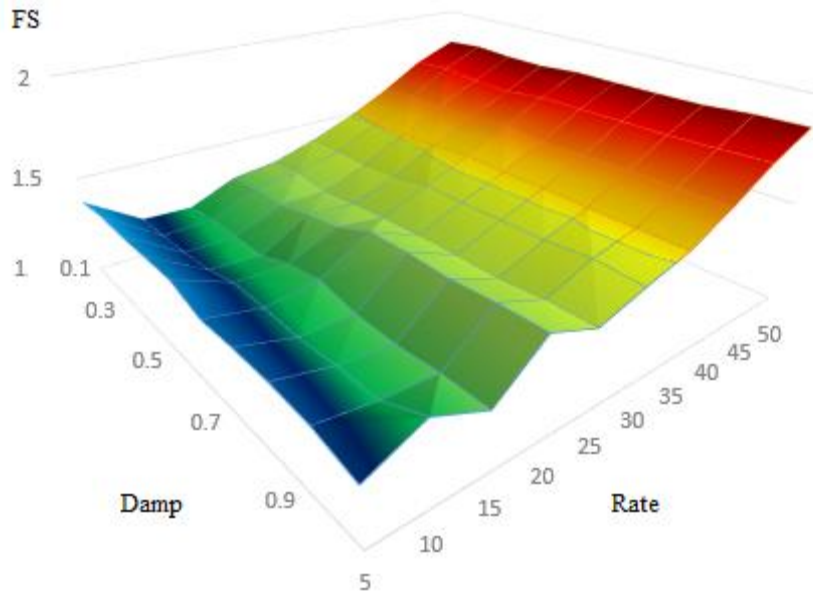

Figure 4: A sample for parameter tuning experiments for CATT method.

## 7-7 Results

Once the parameters are set and the best values are found, it is time to perform the main tests. As mentioned above, ten methods are implemented in this paper. All of these methods are implemented with the best value of their parameters, which is explained in section 7-6. Then, all of these methods are run on the Sep_TD_Tel01 dataset. Topics obtained from each method are evaluated using the metrics of "Topic Precision", "Topic Recall" and "Topic F-measure". These evaluation metrics measure the quality of the obtained topics. Therefore, the results represent how good topics those methods can choose. The results are presented in Table 5, where it can be observed that the CATT method, belonging to the FPCL category, outperforms the other nine methods based on the F-measure metric. In the CL category, the WVOP method and in the FP category, the TSCV method outperforms other methods. Table 6 compares different categories by "Average Topic Precision", "Average Topic Recall" and "Average Topic F-measure". Comparing different categories based on the "Average Topic F-measure", shows the FPCL category outperforms the other two categories. However, this outperformance is not strong as it does not demonstrate superiority in terms of the "Average Topic Precision" metric.

We conducted another experiment to observe the efficiency of the methods in separating and clustering the posts. For this purpose, we extracted the related post of each topic and then evaluated them using the "ClassFS", "ClusterFS", and "MeanFS" metrics. These metrics evaluate the efficiency of the methods in separating posts. Therefore, the results represent the level of success achieved in post separation. The evaluation results are given in Table 7. As shown in the table, the UFPT method, which belongs to the FP category, outperforms the other nine methods according to the MeanFS metric. In the CL category, the WVOP method and in the FPCL category, the SGJP method outperforms other methods. Table 8 compares different categories based on "Average ClassFS", "Average ClusterFS" and "Average MeanFS". Comparing different categories considering the "Average MeanFS", shows the FP category outperforms the other two categories and its superiority is strong, since it outperforms in terms of Average ClassFS and Average ClusterFS.

The UFPT has the best results in Table 7. It not only achieves good results based on the MeanFS metric but also receives positive evaluations from human evaluators. This method produces outputs that are not only has good quality but also easily comprehensible. Its success may be attributed to its focus on the emerging frequency of words across windows, in addition to the frequency of words in the current window. Moreover, the UFPT prioritizes patterns over individual words, making it error-tolerant, meaning that error words do not appear in the output.

Table 5: Comparing different methods implemented in this paper by "Topic Precision", "Topic Recall" and "Topic F-measure" evaluation metrics

| Method | Category | Topic Precision ↑ | Topic Recall ↑ | Topic F-Measure ↑ |
|---|---|---|---|---|
| TSCV | Frequent Pattern (FP) | 0.702703 | 0.413793 | **0.520868** |
| DSFG |  | 0.950820 | 0.224138 | 0.362762 |
| UFPT |  | 0.941176 | 0.172414 | 0.291439 |
| WVOP | Clustering (CL) | 0.931271 | 0.431034 | **0.589310** |
| FTOP |  | 0.942761 | 0.379310 | 0.540968 |
| GLCM |  | 0.818182 | 0.310345 | 0.450000 |
| GLGK |  | 0.714286 | 0.206897 | 0.320856 |
| SGJP | Hybrid (FPCL) | 0.527273 | 0.413793 | 0.463691 |
| CATT |  | 0.516484 | 0.827586 | **0.636030** |
| FHKN |  | 0.908046 | 0.303571 | 0.455023 |

Table 6: Comparing categories by "Average Topic Precision", "Average Topic Recall" and "Average Topic F-measure". The results are the average values of each metric for the methods of that category.

| Category | Method | Avg. Topic Precision ↑ | Avg. Topic Recall ↑ | Avg. Topic F-Measure ↑ |
|---|---|---|---|---|
| Frequent Pattern (FP) | TSCV | 0.864900 | 0.270115 | 0.391690 |
|  | DSFG |  |  |  |
|  | UFPT |  |  |  |
| Clustering (CL) | WVOP | 0.851625 | 0.331897 | 0.475283 |
|  | FTOP |  |  |  |
|  | GLCM |  |  |  |
|  | GLGK |  |  |  |
| Hybrid (FPCL) | SGJP | 0.650601 | 0.514984 | **0.518248** |
|  | CATT |  |  |  |
|  | FHKN |  |  |  |

Table 7: Comparing different methods implemented in this paper by "ClassFS", "ClusterFS" and "MeanFS"

| Method | Category | ClassFS ↓ | ClusterFS ↓ | MeanFS ↓ |
|---|---|---|---|---|
| TSCV | Frequent Pattern (FP) | 0.645277 | 1.594577 | 1.119927 |
| DSFG |  | 0.750816 | 1.402442 | 1.076629 |
| UFPT |  | 0.362058 | 1.729917 | **1.045988** |
| WVOP | Clustering (CL) | 1.219941 | 1.352989 | **1.286465** |
| FTOP |  | 1.141669 | 1.463282 | 1.302475 |
| GLCM |  | 1.028226 | 1.805302 | 1.416764 |
| GLGK |  | 1.234083 | 1.819762 | 1.526923 |
| SGJP | Hybrid (FPCL) | 0.388564 | 1.721218 | **1.054891** |
| CATT |  | 0.738440 | 1.530920 | 1.134680 |
| FHKN |  | 0.825380 | 1.520165 | 1.172773 |

Table 8: Comparing categories by "Average ClassFS", "Average ClusterFS" and "Average MeanFS". The results are the average values of each metric for the methods of that category.

| Category | Method | Avg. Class FS ↓ | Avg. Cluster FS ↓ | Avg. Mean FS ↓ |
|---|---|---|---|---|
| **Frequent Pattern (FP)** | TSCV | 0.586050 | 1.575645 | **1.080848** |
| | DSFG | | | |
| | UFPT | | | |
| **Clustering (CL)** | WVOP | 1.155980 | 1.610334 | 1.383157 |
| | FTOP | | | |
| | GLCM | | | |
| | GLGK | | | |
| **Hybrid (FPCL)** | SGJP | 0.650795 | 1.590768 | 1.120781 |
| | CATT | | | |
| | FHKN | | | |

## 7-8 Discussion

As mentioned previously, if we consider the average topic F-measure (Table 5), FPCL methods are better than other categories. However, in terms of the average MeanFS (Table 7), the FP category methods are better than the others. The FS focuses on posts related to a single topic, while the topic F-measure focuses on the words in the topics. Therefore, it can be concluded that if the goal is to separate posts related to a single topic (e.g., with the aim of further investigations), it is better to use FP methods. However, if the goal is to achieve the right sequence of keywords that give a user-friendly topic, it is better to use FPCL methods.

It is worth considering the following points about the advantages and disadvantages of each category of methods. FP category methods is based on the FP-Growth algorithm. This algorithm's major problem is determining the value of min support. Allocating a high value to this parameter will weaken the results and lead to the loss of some topics. In contrast, allocating a low value increases the consumed time and memory of the algorithm even more than system memory, making it impossible to run. Manually determining the optimal value for min support is practically impossible, since each window's value differs from the other windows. Therefore, in practice, two approaches are used: First, a constant value is considered for all windows, which is not an efficient approach. Second, some heuristic methods are proposed and used to determine the value of min support in each window automatically. But these heuristic methods are also not very efficient. We conducted experiments for this purpose during this research. We set some windows' optimal value of min support by trial and error. It was observed that the determined values by heuristic methods are far from this optimal value in many windows. As a result, it can be said that determining the value of min support is the main disadvantage of FP category methods. In contrast, these methods have an important advantage which is they are faster than CL and FPCL category methods.

The advantage of CL category methods lies in their ease of implementation. However, their implementation relies on a suitable embedding model. One limitation faced during this research was that for many embedding methods, there was no suitable pre-trained embedding model for Persian, which is trained with high vocabulary size. In other words, when we used the existing pre-trained models, we often encountered a high out-of-vocab rate. Therefore, we couldn't use some of the very new embedding methods and we had to choose the methods which has suitable pretrained models. Even though CL methods are easy to implement, determining the number of clusters is their major drawback. The number of clusters in these methods is equal to the number of topics. Therefore, it is significant to determine the accurate number of clusters. In methods such as k-means, the k parameter determines the number of clusters. In methods such as DBSCAN and OPTICS, parameters like min-points and neighborhood radius determine the number of clusters. Since the number of topics in each window is unknown, the number of clusters in each window cannot be determined in advance. Therefore, the algorithm should be run for different values of parameter, then the optimal value of the parameters is determined using metrics such as Silhouette, which is time consuming. Graph-based clustering methods use the majority metric to determine the number of clusters in the same process [36], which is also time consuming. Sometimes, despite a lot of time spent to determine the correct number of clusters, the determined value is different from the actual number of clusters, which is another problem in these methods. Therefore, it can be said that the ease of implementation is the top advantage of the CL category, and determining the number of clusters is its main disadvantage. (Determining the number of clusters is not only time consuming but also hard to reach optimum value.)

FPCL category methods are a combination of FP and CL categories and, therefore, they almost have their advantages and disadvantages. However, there are some differences since they are not an exact combination of FP and CL. The FHKN method is the most exact combination of FP and CL in the FPCL category. It utilizes the combination of the FP-Growth and HUP algorithms to discover frequent patterns and applies the combination of k-means and Newman for clustering. This method is faster than CL methods because instead of clustering on posts, it performs clustering on frequent patterns that are smaller in size than posts. Additionally, the speed of this method is significantly higher than the other two methods in the FPCL category. The results of the evaluation of its output by humans show that this method is successful in finding topics that have a large number of posts and its output is more understandable for humans. In contrast, the CATT method is successful in finding topics with a mean number of posts. However, the sequence of words it generates for the topics is not very user-friendly. The SGJP method succeeds in clustering posts, but it is slower because it uses graph-based clustering, which is time-consuming. This method has more parameters to tune, which also takes time to setup. Additionally, extracting the topic word sequence requires background knowledge, which is laborious and time consuming to provide. In this research, we extracted this background knowledge from Persian Wikipedia papers. More than 3.2 million Persian Wikipedia documents were processed for this purpose. It took over 40 days to process this number of documents and extract the necessary background knowledge. Although this knowledge only needs to be extracted once during the system setup and does not affect processing time, it is still considered one of the disadvantages of this method.

# 8   Conclusion

As mentioned above, the basis of this paper started from the fact that we needed to implement a topic detector for the Persian language for an application. However, since Persian is a low-resource language, a few papers have been found in Persian. Therefore, we had to implement 10 methods inorder to choose one that produces acceptable result in Persian. Consequently, all the methods in this paper have been re-implemented for the Persian language with some modifications to the original English language methods. In other words, we wrote the code for these methods from scratch ourselves.

There are several methods for topic detection, which are classified into different categories. In this research, the FP and CL categories are selected, and a new category called FPCL, which combinations these two categories, is proposed. Then 10 methods from these three categories are choosen and implemented. Those include three methods, TSVC, DSFG, and UFPT, from the FP category; four methods, WVOP, FTOP, GLCM, and GLGK, from the CL category; and three methods, SGJP, CATT, and FHKN, from the FPCL category. A large volume of datasets is used, including Hamshahri with approximately 67 million tokens, Telegram with around 100 million tokens and Wikipedia with approximately 712 million tokens. In total, approximately 1.4 billion tokens are processed to provide background knowledge.

In addition to the well-known metrics: "Topic Precision", "Topic Recall", and "Topic F-measure", newly introduced metrics: "ClassFS", "ClusterFS", and "MeanFS" are used to evaluate the methods. The FS is a new multiclass - multicluster metric which is used for the first time in topic detection in this paper. Additionally, the silhouette evaluation metric is employed for parameter tuning. The results indicate that considering the topic F-measure, which measures the quality of selected terms for the topic, the CATT method outperforms other methods. However, when considering MeanFS, which measures post separation quality, the UFPT method outperforms.

Comparing these three categories based on the experimental results, if the aim is to extract topics for ongoing posts in the data stream, FPCL category methods are recommended as they usually generate better topics. However, if the aim is to cluster the posts and separate the related posts (e.g., for later investigation by the user), then the FP category methods outperform. Methods that use only clustering (CL category) usually do not have good results in comparison with other two categories.